\documentclass{tlp}
\usepackage{mathptmx}
\usepackage{comment}
\long\def\comment#1{}

\usepackage[linesnumbered,ruled,slide,boxed,vlined]{algorithm2e}

\usepackage{amssymb}
\usepackage{amsmath}

\usepackage[english]{babel}

\usepackage{verbatim}

\newtheorem{theorem}{Theorem}[section]
\newtheorem{proposition}{Proposition}[section]
\newtheorem{example}{Example}
\newtheorem{lemma}{Lemma}[section]
\newtheorem{corollary}[theorem]{Corollary}
\newtheorem{definition}{Definition}[section]

\newcommand{\imply}{\supset}

\newcommand\ie{{\it i.e.}}
\newcommand\eg{{\it e.g.} }
\newcommand\st{{\it s.t.} }
\newcommand\wrt{{\it w.r.t.\@} }

\newcommand{\MKNF}{\textmd{\rm MKNF} }

\newcommand{\Not}{{\textit {\bf  not}\,}}
\newcommand{\bfnot}{{\textit {\bf  not}}}
\newcommand{\K}{\textit {\bf  K}\,}
\newcommand{\bfK}{{\textit {\bf  K}}}

\newcommand{\head}{\textit{head}}
\newcommand{\body}{\textit{body}}

\newcommand{\atmost}{\textrm \sf Atmost}

\newcommand{\KA}{\sf KA}
\newcommand{\OB}{\sf OB}

\newcommand{\boldP}{\textit {\bf  P}}
\newcommand{\boldN}{\textit {\bf  N}}

\newcommand{\UP}{\textit{UP}}

\newcommand{\cK}{\mathcal K}
\newcommand{\cO}{\mathcal O}
\newcommand{\cP}{\mathcal P}
\newcommand{\cL}{\mathcal L}
\newcommand{\WFM}{\textit{WFM}}

\title[Well-Founded Operators for Normal Hybrid MKNF Knowledge Bases]{Well-Founded Operators for Normal Hybrid MKNF Knowledge Bases}

\author[J. Ji, F. Liu, J. You]
         {Jianmin Ji$^1$, Fangfang Liu$^2$, Jia-Huai You$^3$
         \\
          $^1$University of Science and Technology of China, Hefei, China\\
          $^2$Shanghai University, Shanghai, China\\
          $^3$University of Alberta, Edmonton, Canada
         }

\begin{document}

\maketitle

\begin{abstract}
Hybrid \MKNF knowledge bases have been considered one of the dominant approaches to combining open world ontology languages with closed world rule-based languages. Currently, the only known inference methods are based on the approach of guess-and-verify, while most modern SAT/ASP solvers are built under the DPLL architecture.
The central impediment here is that it is not clear what constitutes a constraint propagator, a key component employed in any DPLL-based solver.
In this paper, we address this problem by formulating the notion of unfounded sets for nondisjunctive hybrid \MKNF knowledge bases, based on which we propose and study two new well-founded operators.  
We show that by employing a well-founded operator as a constraint propagator, a sound and complete DPLL search engine can be readily defined. We compare our approach with {the operator based on} the alternating fixpoint construction by Knorr et al [2011] and show that, when applied to arbitrary partial partitions,  the new well-founded operators not only propagate more truth values but also circumvent the non-converging behavior of the latter.  
In addition, we study the possibility of simplifying a given hybrid \MKNF knowledge base by employing a well-founded operator and show that, out of the two operators proposed in this paper, the weaker one
can be applied for this purpose  and the stronger one cannot. 
These observations are useful in implementing a grounder for  hybrid \MKNF knowledge bases, {which can be applied} before  the computation of \MKNF models.

The paper is under consideration for acceptance in TPLP.

\end{abstract}

  \begin{keywords}
    Hybrid \MKNF\!, Constraint propagation, DPLL-based search engine, Well-founded operator.
  \end{keywords}


\section{Introduction}

Hybrid \MKNF knowledge bases~\cite{motik2010reconciling}, based on the logic of minimal knowledge and negation as failure (MKNF)~\cite{lifschitz1991nonmonotonic}, is one of the most influential yet mature formalisms for combining open world ontology languages, such as description logics (DLs)~\cite{baader2003description} and the OWL-based ones~\cite{hitzler2009owl}, with closed world rule-based languages, like logic programs under the stable model semantics~\cite{baral2003knowledge}. 
The semantics of hybrid \MKNF knowledge bases is captured by \MKNF models.
It is shown that the data complexity of reasoning within hybrid \MKNF knowledge bases is in many cases not higher than reasoning in the corresponding fragment of logic programming~\cite{motik2010reconciling}. For instance, if the underlying DL fragment is of polynomial data complexity, then the data complexity of instance checking after combining with nondisjunctive (normal) rules is coNP-complete. However, despite many efficient solvers for logic programs~\cite{heule2015s}, there is few work on computing \MKNF models of hybrid \MKNF knowledge bases---the only known reasoning methods are based on the brute-force, guess-and-verify approach \cite{motik2010reconciling}. In this approach, the set of $\bfK$-atoms is partitioned into two subsets, the set of true $\bfK$-atoms and the set of false $\bfK$-atoms, in each possible way, and whether it corresponds to an \MKNF model is verified by an operator similar to the {immediate consequence operator} in logic programming.  

Most modern SAT/ASP solvers are built under the DPLL architecture \cite{SMT-JACM}, where propagating a partial assignment is a key process. Recall that
 a DPLL-based solver is a search engine whose basic operation is to make decisions, 
propagate a partial assignment at each decision point, and backtrack when {a conflict is encountered. 
 Typically, a competitive solver also implements powerful heuristics for variable selection, and conflict analysis and clause learning \cite{DBLP:conf/cade/ZhangM02}. }
Propagating a partial assignment can result in substantial pruning of the search space---all the propagated truth values are committed 
in expanding the given partial assignment. In this context, the larger the computed set of truth values, the stronger is the propagator. Apparently, the cost of computing such a set should also be taken into consideration. 
As an example, BCP (Boolean Constraint Propagation, also called Unit Propagation) is considered the most important part of a SAT solver \cite{DBLP:journals/cacm/MalikZ09}, and a SAT solver typically  spends more than $80\%$ of its time running BCP.
In ASP, the well-known {\em Expand} function in {\tt smodels} \cite{DBLP:journals/ai/SimonsNS02} plays a central role in constraint propagation for weight constraint logic programs, but the feature of {\em lookahead} is often abandoned due to its high cost. {Also,} viewing inferences in ASP as unit propagation on nogoods, along with other techniques, has made {\tt clasp} among the most competitive solvers for ASP as well as for SAT \cite{DBLP:journals/ai/GebserKS12}. More recently, for answer set programs with external sources \cite{DBLP:conf/ijcai/EiterKRW16}, the approach of guessing  truth values of external sources is replaced with evaluations under partial assignments, which produces substantial gains in search efficiency. 

Despite all of these advances, for hybrid \MKNF knowledge bases, the fundamental issue of what constitutes constraint propagation for a DPLL-based  search engine has not been addressed.  The brute-force, guess-and-verify proof method is still the state-of-the-art. 

To formulate a well-founded semantics for normal hybrid \MKNF knowledge bases, 
\citeANP{knorr2011local}~\shortcite{knorr2011local} proposed 
a well-founded operator to compute consequences that are satisfied by every \MKNF model of a hybrid \MKNF knowledge base by an alternating fixpoint construction. The operator computes the least fixpoint iteratively from the least element in a bilattice and enjoys a polynomial data complexity when the underlying DL is polynomial.  

It is important to distinguish constraint propagation from 
computing the well-founded semantics \-- while the latter computes one least fixpoint, the former can be viewed as computations by
a family of operators, each of which is applied to a different partial partition (partial partitions are analogue to partial interpretations in SAT/ASP). We say that such an operator is {\em instantiated}, or {\em induced}, from the related partial partition, and call it an {\em instance operator}. If such an
instance operator is monotonic, we then can analyze its properties by applying the Knaster-Tarski fixpoint theory \cite{Tarski1955} and view the computation of its least fixpoint as the process of constraint propagation that extends the given partial partition.  Thus, in this paper the term {\em well-founded operator} refers to the corresponding family of instance operators.  We show that if we  apply this idea to Knorr et al.'s operator, an instance operator may not be converging. Thus, Knorr et al.'s operator does not provide a satisfactory solution for constraint propagation. 

In this paper, we address the problem of constraint propagation for 
normal hybrid \MKNF knowledge bases.
The main contributions are the following:
\begin{itemize}
  \item We extend the notion of unfounded sets to normal hybrid \MKNF knowledge bases and show that desirable properties for logic programs~\cite{leone1997disjunctive} can be generalized to normal
hybrid \MKNF knowledge bases; in particular, \MKNF models are precisely unfounded-free models.  We provide a procedure to compute the greatest unfounded set of a normal hybrid \MKNF knowledge base \wrt a partial partition, which has polynomial data complexity when the underlying DL is polynomial.
  \item We introduce two new well-founded operators, with one being stronger than the other. 
We show that both are stronger than the one proposed in~\cite{knorr2011local} when applied to arbitrary partitions.

\item  Employing either of the two operators as the underlying propagator, we formulate a DPLL-based procedure to determine whether an \MKNF model exists for a normal hybrid \MKNF knowledge base; {in case the answer is positive, the procedure can be adopted to compute all \MKNF models by backtracking. }
This provides another DPLL-based NP inference engine, as  the decision problem is NP-complete when the underlying DL component is trackable \cite{motik2010reconciling}. 

\item We show that the two proposed operators have different utilities. The stronger one serves as a stronger propagator in a DPLL-based search engine, and the weaker one has the desired property that it
 can be used to simplify the given hybrid \MKNF knowledge base before we proceed to compute  \MKNF models.  It thus provides a theoretical basis for implementing the simplification process in a grounder for {normal} hybrid \MKNF knowledge bases. 


\end{itemize}

The paper is completed with related work, followed by conclusions and future directions. The proofs are moved to \ref{proof-appendix}, with \ref{unfounded-compare} providing a detailed comparison with the notion of unfounded set mentioned in a proof in \cite{knorr2011local}.

\section{Preliminaries}

\subsection{Minimal knowledge and negation as failure}

The logic of minimal knowledge and negation as failure (MKNF)~\cite{lifschitz1991nonmonotonic} is based on a first-order language~$\cL$ (possibly with equality $\approx$) with two modal operators, $\K$\!, for minimal knowledge, and $\Not$\!, for negation as failure.
In \MKNF\!\!, a {\em first-order atom} is a formula of the form $P(t_1, \ldots, t_n)$, where $t_i$ are terms and $P$ is a predicate in $\cL$.
 {\em \MKNF  formulas} are first-order formulas with $\K$ and $\Not$. 
An \MKNF formula $F$ is {\em ground} if it contains no variables, and $F[t/x]$ is the formula obtained from $F$ by replacing all free occurrences of the variable $x$ with term $t$.

A {\em first-order interpretation} is understood as in first-order logic. The universe of a first-order interpretation~$I$ is denoted by $\left|I\right|$.
A {\em first-order structure} is a nonempty set $M$ of first-order interpretations with the universe $\left|I\right|$ for some fixed $I\in M$.
An {\em \MKNF structure} is a triple $(I, M, N)$, where $M$ and $N$ are sets of first-order interpretations with the universe $\left|I\right|$. We define the {\em satisfaction relation} $\models$ between an \MKNF structure $(I, M, N)$ and an \MKNF formula $F$. Then we extend the language $\cL$ by object constants representing all elements of $\left|I\right|$ and call these constants {\em names}:
\begin{itemize}
\item $(I, M, N)\models A$ ($A$ is a first-order atom) if $A$ is true in~$I$,
\item $(I, M, N)\models \neg F$ if $(I, M, N)\not\models F$,
\item $(I, M, N)\models F\land G$ if $(I, M, N)\models F$ and $(I, M, N)\models G$,
\item $(I, M, N) \models \exists x F$ if $(I, M, N) \models F[\alpha/x]$ for some name $\alpha$,
\item $(I, M, N) \models \K F$ if $(J, M, N) \models F$ for all $J\in M$,
\item $(I, M, N)\models \Not F$ if $(J, M, N)\not\models F$ for some $J\in N$.
\end{itemize}
The symbols $\top$, $\bot$, $\lor$, $\forall$, and $\supset$ are interpreted as usual.

\begin{comment}
From the definition, following \MKNF formulas are satisfied by any \MKNF structure:
\begin{align*}
  \K(\forall x\, \varphi) &\equiv \forall x\, \K \varphi, &
  \Not(\forall x\, \varphi) &\equiv \forall x\, \Not \varphi.
\end{align*}
Hence, we can replace any subformula of an \MKNF formula of the form on the left of $\equiv$ with the formula on the right and vice-versa.
An \MKNF formula is in {\em prenex form} if it is written as $Q_1 x_1 \cdots Q_n x_n F$ where each $Q_i \in \{\exists, \forall\}$ and $F$, called {\em matrix}, is  quantifier-free. Every \MKNF formula can be equivalently translated to an \MKNF formula in prenex form in the sense of \MKNF structures.
\end{comment}

An {\em \MKNF interpretation} $M$ is a nonempty set of first-order interpretations over the universe $\left|I\right|$ for some $I\in M$.
An \MKNF interpretation $M$ {\em satisfies} an \MKNF formula $F$, written $M\models_\MKNF F$, if $(I, M, M)\models F$ for each $I\in M$.
\begin{definition}
An \MKNF interpretation $M$ is an {\em \MKNF model} of an \MKNF formula~$F$ if
\begin{enumerate}
  \item $M\models_\MKNF F$, and
  \item there is no \MKNF interpretation $M'$ such that $M'\supset M$ and $(I', M', M)\models F$ for every $I'\in M'$.
\end{enumerate}
\end{definition}

For example, with the \MKNF formula $F = \Not b\imply \K a$, it is easy to verify that the \MKNF interpretation $M = \{ \{a\}, \{a, b\}\}$ is an \MKNF model of $F$.

In this paper, we consider only \MKNF formulas that do not contain nested occurrences of modal operators and every first-order atom {occurring}  in the formula is in the range of a modal operator. 
Specifically, a {\em $\bfK$-atom} is a formula of the form $\K \psi$ and a {\em $\bfnot$-atom} is a formula of the form $\Not \psi$, where $\psi$ is a first-order formula.

\subsection{Hybrid \MKNF knowledge bases}

Following \cite{motik2010reconciling}, a {\em hybrid \MKNF knowledge base} ${\cK} = ({\cO}, {\cP})$ consists of a decidable description logic (DL) knowledge base ${\cO}$ translated into first-order logic and a {\em rule base} $\cP$, which is a finite set of {\em \MKNF rules}. An \MKNF rule~$r$ has the following form, where $0\leq k\leq m\leq n$, and $a_i$ are function-free first-order atoms:
\begin{equation}\label{eq:rule}
\K a_1 \lor \ldots \lor \K a_k \gets \K a_{k+1}, \ldots, \K a_{m}, \Not a_{m+1}, \ldots, \Not a_n. 
\end{equation}
If $k= 1$, $r$ is a {\em normal \MKNF rule}; if $m=0$, $r$ is a {\em positive \MKNF rule}; if $k=1$ and $n=m=0$, $r$ is an {\em \MKNF fact}. A {\em hybrid \MKNF knowledge base} ${\cK}=({\cO}, {\cP})$ is {\em normal} if all \MKNF rules in $\cP$ are normal; $r$ is {\em ground} if it does not contain variables; and 
$\cP$ is {\em ground} if all \MKNF rules in $\cP$ are ground.

We also write an \MKNF rule $r$ of form~\eqref{eq:rule} as $\head(r)\gets \body(r)$, where $\head(r)$ is $\K a_1 \lor \cdots \lor \K a_k$, $\body(r) = \body^+(r) \land \body^-(r)$, $\body^+(r)$ is $\K a_{k+1}\land \cdots \land \K a_m$, and $\body^-(r)$ is $\Not a_{m+1} \land \cdots \land \Not a_n$, and we identify $\head(r)$, $\body(r)$, $\body^+(r)$, $\body^-(r)$ with their corresponding sets of $\bfK$-atoms and $\bfnot$-atoms. With a slight abuse of notion, we denote $\K (body^-(r)) = \{ \K a\mid \Not a\in body^-(r)\}$.

Let ${\cK} = ({\cO}, {\cP})$ be a hybrid \MKNF knowledge base and $r$ an \MKNF rule. 
We define an operator $\pi$ for $r$, $\cP$, $\cO$ and $\cK$, respectively, as follows, where $\vec{x}$ is the vector of the free variables appearing in $r$:
\begin{align*}
\pi(r) &= \forall \vec{x}.\, (\body(r)\imply \head(r) ),\\
\pi({\cP}) &= \bigwedge_{r\in {\cP}} \pi(r),\\
\pi({\cO}) & \text{ is a corresponding function-free first-order logic formula},\\
\pi({\cK}) &= \K \pi({\cO}) \land \pi({\cP}).
\end{align*}
For simplicity, in the rest of this paper we may identify $\cK$ with the \MKNF formula $\pi({\cK})$.

An \MKNF rule $r$ is {\em DL-safe} if every variable in $r$ occurs in at least one non-DL-atom $\K a$ occurring in the body of $r$. A hybrid \MKNF knowledge base $\cK$ is {\em DL-safe} if all \MKNF rules in $\cK$ are DL-safe.
A notion called {\em standard name assumption} is applied to hybrid \MKNF knowledge bases to avoid unintended behavior~\cite{motik2010reconciling}, under which interpretations are Herbrand ones with a countably infinite number of additional constants.
If ${\cK}$ is DL-safe, then ${\cK}$ is semantically equivalent to ${\cK}' = ({\cO}, {\cP}')$ in terms of \MKNF models where ${\cP}'$ is ground, hence decidability is guaranteed.

In the rest of this paper, we consider normal hybrid \MKNF knowledge bases containing ground \MKNF rules and use the standard name assumption for first-order inferences.

\subsection{Alternating fixpoint construction}

We briefly review the operator based on an alternating fixpoint construction introduced in \cite{knorr2011local}.

Let ${\cK} = ({\cO}, {\cP})$ be a (ground) hybrid \MKNF knowledge base. The set of $\bfK$-atoms of $\cK$, written ${\KA}({\cK})$, is the smallest set that contains:
\begin{enumerate}
  \item all ground $\bfK$-atoms occurring in $\cP$, and
  \item a $\bfK$-atom $\K a$ for each ground $\Not\!$-atom $\Not a$ occurring in $\cP$.
\end{enumerate}
A {\em partial partition} $(T, F)$ of ${\KA}({\cK})$ consists of two sets, where $T,\, F\subseteq {\KA}({\cK})$ and $T\cap F = \emptyset$.
For a subset $S$ of ${\KA}({\cK})$, the {\em objective knowledge} of $S$ \wrt$\!{\cK}$ is the set of first-order formulas $ {\OB}_{{\cO},\, S} = \{\pi({\cO})\}\cup \{ a\mid \K a\in S\}$.

For two pairs $(T_1, F_1)$ and $(T_2, F_2)$, we define $(T_1, F_1) \sqsubseteq (T_2, F_2)$ if $T_1\subseteq T_2$ and $F_1\subseteq F_2$, $(T_1, F_1) \sqsubset (T_2, F_2)$ if $(T_1, F_1)\sqsubseteq (T_2, F_2)$ and $(T_1, F_1) \neq (T_2, F_2)$, and $(T_1, F_1) \sqcup (T_2, F_2) = (T_1\cup T_2, F_1\cup F_2)$.

Let ${\cK}= ({\cO}, {\cP})$ be a normal hybrid \MKNF knowledge base and $S\subseteq {\KA}({\cK})$. The operators $T^*_{{\cK},\, S}$, $T^{*\prime}_{{\cK},\, S}$ are defined on subsets of ${\KA}({\cK})$ as follows:
\begin{align*}
T^*_{{\cK},\, S}(X) = &\ \{ \K a \mid r\in {\cP},\, \K a\in \head(r),\, \body^+(r)\subseteq X, \K(\body^-(r))\cap S=\emptyset \} \ \\\ &\ \cup \{\K a\in {\KA}({\cK}) \mid  {\OB}_{{\cO},\, X}\models a\}, \\
T^{*\prime}_{{\cK},\, S}(X) =&\ \{\K a\mid r\in {\cP},\, \K a\in \head(r),\, \body^+(r)\subseteq X, \K(\body^-(r))\cap S=\emptyset,\\ &\ \text{ and $ {\OB}_{{\cO},\, S}\cup \{a\}$ is consistent} \} \cup \{\K a \mid \K a\in {\KA}({\cK}),\,  {\OB}_{{\cO},\, X}\models a\}.
\end{align*}

Note that, both $T^*_{{\cK},\, S}$ and $T^{*\prime}_{{\cK},\, S}$ are monotonic. We denote by  $\Gamma_{{\cK}}(S)$ and $\Gamma'_{{\cK}}(S)$, respectively, the least fixpoint of the corresponding operator.

Let ${\cK}$ be a normal hybrid \MKNF knowledge base. 
We define two sequences ${\boldP}_i$ and ${\boldN}_i$ as follows:
\begin{align*}
{\boldP}_0 &= \emptyset, & {\boldN}_0 &= {\KA}({\cK}),\\
{\boldP}_{n+1} &= \Gamma_{{\cK}}({\boldN}_n), & \boldN_{n+1} &= \Gamma'_{{\cK}}({\boldP}_n),\\
{\boldP}_\omega &= \bigcup {\boldP}_i, & {\boldN}_\omega &= \bigcap {\boldN}_i.
\end{align*}

\begin{definition}
Let ${\cK}$ be a normal hybrid \MKNF knowledge base. 
The {\em coherent well-founded partition} 
of  ${\cK}$ is defined by $({\boldP}_\omega,\, {\KA}({\cK})\setminus {\boldN}_\omega)$.\footnote{{Note that, in general, $({\boldP}_\omega,\, {\KA}({\cK})\setminus {\boldN}_\omega)$ may not be consistent, i.e., it is not guaranteed that the condition ${\boldP}_\omega \cap {\KA}({\cK})\setminus {\boldN}_\omega \not = \emptyset$ holds.}}
\end{definition}

Clearly,  the number of iterations in the construction of the  coherent well-founded partition is linear in the number of $\bfK$-atoms in ${\KA}({\cK})$. If the entailment relation $ {\OB}_{{\cO},\, S}\models a$ can be computed in polynomial time, so can each iteration as well as the coherent well-founded partition.


\section{Unfounded Set and Well-Founded Operators}

In this section, we define the notion of unfounded set for (ground) normal hybrid \MKNF knowledge bases, {present an algorithm to compute the greatest unfounded set, and then 
introduce two new well-founded operators. At the end, we} discuss the relations of these operators with the one based on the alternating fixpoint construction. 

\subsection{Unfounded sets}

{In logic programming, an unfounded set in general refers to a set of atoms that fail to be derived by rules. In the context of hybrid \MKNF knowledge bases, the concept becomes more involved due to possible inferences with the knowledge expressed in the underlying ontology.}

Given a set of normal \MKNF rules $R$, we define $\head(R) = \{ a \mid {\K a}\in \head(r)\text{ for some $r\in R$} \}$.

\begin{definition}\label{unfounded}
A set $X\subseteq {\KA}({\cK})$ is an {\em unfounded set} of a normal hybrid \MKNF knowledge base ${\cK}=({\cO}, {\cP})$ \wrt a partial partition $(T, F)$ of ${\KA}({\cK})$, if for each $\K a\in X$ and each $R\subseteq {\cP}$ such that 
\begin{itemize}
  \item $\head(R)\cup  {\OB}_{{\cO}, T} \models a$, and
  \item  for each $\K b\in F$, $\head(R)\cup  {\OB}_{{\cO}, T} \cup \{\neg b\}$ is consistent, in particular, $\head(R)\cup  {\OB}_{{\cO}, T}$ is consistent when $F=\emptyset$,
\end{itemize}
there exists an \MKNF rule $r\in R$ satisfying one of the following conditions:
\begin{itemize}
  \item $\body^+(r) \cap F\neq \emptyset$,
  \item $\K(\body^-(r))\cap T\neq\emptyset$, or
  \item $\body^+(r)\cap X\neq\emptyset$.
\end{itemize}
A $\bfK$-atom in an unfounded set is called an {\em unfounded atom}. 
\end{definition}

\begin{comment}
Let ${\cK} = ({\cO}, {\cP})$ be a normal hybrid \MKNF knowledge base, $(T, F)$ a partial partition of ${\KA}({\cK})$.
A set $X$ of $\bfK$-atoms in ${\KA}({\cK})$ is an {\em unfounded set} of ${\cK}$ \wrt $(T, F)$ if for each $\bfK$-atom $\K a\in X$ and each set of \MKNF rules $R\subseteq {\cP}$ with $\head(R)\cup  {\OB}_{{\cO}, T} \models a$ and $\head(R)\cup  {\OB}_{{\cO}, T} \cup \{\neg b\}$ is consistent for each $\K b\in F$, there exists an \MKNF rule $r\in R$ satisfying one of the following conditions:
\begin{itemize}
  \item $\body^+(r) \cap F\neq \emptyset$,
  \item $\K(\body^-(r))\cap T\neq\emptyset$, or
  \item $\body^+(r)\cap X\neq\emptyset$.
\end{itemize}
\end{comment}

Roughly speaking, for a modal atom $\K a \in X$ to be unfounded \wrt $(T,F)$, any group of rules $R$ that can help derive it, along with ${\OB}_{{\cO}, T}$,
 must contain at least one rule which is not applicable given $(T,F)$. Since the condition must be satisfied for each $R$, when $R$ is a minimal set such that 
$\head(R)\cup  {\OB}_{{\cO}, T} \models a$, the existence of such a rule blocks the derivation.

More precisely,
an unfounded set $X$ \wrt $(T,F)$ is one  such that for each $\K a \in X$, if $a$ is derivable from (the objective heads of) rules in $R$ and objective knowledge 
${\OB}_{{\cO}, T}$, where ${\OB}_{{\cO}, T}$ is not in conflict with any false atom based on $F$, then there exists at least one rule in $R$ such that either its body is not satisfied by $(T,F)$ or the body being satisfied depends on some atoms in $X$.

It is not difficult to verify that, when ${\cO}=\emptyset$, this notion of unfounded sets coincides with the one  for the corresponding logic programs \cite{van1991well}.

\begin{example}\label{exp:1}
Consider ${\cK}_1 = ({\cO}_1, {\cP}_1)$, where $\pi({\cO}_1) = \neg c$ and ${\cP}_1 = \{
\K a \gets \Not b.  ~\K b  \gets \Not a.  ~\K c \gets \K a.\}$.
Since there exists no $R\subseteq {\cP}_1$ with $\head(R)\cup  {\OB}_{{\cO}_1, \emptyset}\models c$ and $head(R)\cup {\OB}_{{\cO}_1, \emptyset}$ is consistent, 
$\{\K c\}$ is an unfounded set of ${\cK}_1$ \wrt $(\emptyset, \emptyset)$.
\end{example}

\begin{proposition}
\label{union}
Let ${\cK}$ be a normal hybrid \MKNF knowledge base, $(T, F)$ a partial partition of ${\KA}({\cK})$. If $X_1$ and $X_2$ are unfounded sets of ${\cK}$ \wrt $(T, F)$, then $X_1\cup X_2$ is an unfounded set of ${\cK}$ \wrt $(T, F)$.
\end{proposition}

As the union of two unfounded sets is also an unfounded set,  the {\em greatest unfounded set} of ${\cK}$ \wrt $(T, F)$, denoted $U_{{\cK}}(T, F)$, exists,
which is the union of all unfounded sets of ${\cK}$ \wrt~$(T, F)$.

\begin{proposition}\label{prop:2}
Let ${\cK}$ be a normal hybrid \MKNF knowledge base, $(T, F)$ a partial partition of ${\KA}({\cK})$, and $U$ an unfounded set of ${\cK}$ \wrt $(T, F)$. For any \MKNF model $M$ of ${\cK}$ with $M\models_\MKNF \bigwedge_{\K a\in T} \K a \land \bigwedge_{\K b\in F} \neg \K b$, $M\models_\MKNF \neg \K u$ for each $\K u\in U$.
\end{proposition}

In logic programming, a declarative characterization of stable models is that they are precisely {\em unfounded-free} models (see, \eg \cite{Alviano11}). The same property holds for normal hybrid 
\MKNF knowledge bases under the notion of unfounded set defined in this paper. 

\begin{proposition}\label{unfounded-free}
Let ${\cK} = ({\cO}, {\cP})$ be a normal hybrid knowledge base and $M$ an \MKNF model of $\cK$. Define $(T,F)$ by $T = \{\K a \in {\KA}({\cK}) \mid M \models_\MKNF \K a\}$ and  $F = {\KA}({\cK})\setminus T$.  Then, $F$ is the greatest unfounded set of 
${\cK}$  \wrt $(T,F)$.
\end{proposition}

We provide an approach to computing the greatest unfounded set of ${\cK}$ \wrt $(T, F)$, \ie, $U_{{\cK}}(T, F)$.
First, we define an operator $V_{{\cK}}^{(T, F)}$ as follows:
\begin{multline*}
V_{{\cK}}^{(T, F)}(X) = \{\K a\in {\KA}({\cK}) \mid {\OB}_{{\cO},\, X}\models a\} \\
\cup \{ \K a\mid r\in {\cP},\, \K a\in \head(r),\, \body^+(r)\subseteq X,\, \body^+(r)\cap F=\emptyset,\, \K(\body^-(r)) \cap T=\emptyset,\\\text{ and $\{a, \neg b\}\cup  {\OB}_{{\cO}, T}$ is consistent for each $\K b\in F$}\}.
\end{multline*}

Clearly, $V_{{\cK}}^{(T, F)}$ is monotonic. We thus define the function ${\atmost}_{{\cK}}(T, F)$ to be the least fixpoint of $V_{{\cK}}^{(T, F)}$. We show that the greatest unfounded set can be computed from ${\atmost}_{{\cK}}(T, F)$.

\begin{theorem}
\label{theorem1}
Let ${\cK}$ be a normal hybrid \MKNF knowledge base and $(T, F)$ a partial partition of ${\KA}({\cK})$. $U_{{\cK}}(T, F) = {\KA}({\cK}) \setminus {\atmost}_{{\cK}}(T, F)$.
\end{theorem}

Clearly, the number of iterations in the construction of ${\atmost}_{\cK}(T, F)$ is linear in the number of $\bfK$-atoms in ${\KA}(\cK)$.
If the entailment relation $ {\OB}_{{\cO},\, X}\models a$ can be computed in polynomial time, then $V^{(T, F)}_{\cK}(X)$ can be computed in polynomial time, and the same holds for computing the greatest unfounded set of $\cK$ \wrt $(T, F)$.




\subsection{A well-founded operator}

{By applying the process of computing the greatest unfounded set \wrt a partial partition, we can define a new well-founded operator. }

Let ${\cK} = ({\cO}, {\cP})$ be a normal hybrid \MKNF knowledge base, and $(T, F)$ a partial partition of ${\KA}({\cK})$. We introduce the well-founded operator $W_{{\cK}}^{(T, F)}$ of ${\cK}$ as follows:
\begin{comment}
\begin{align*}
T_{{\cK}}(T, F) =&\ \{ \K a \mid r\in {\cP},\, \K a\in \head(r),\, \body^+(r)\subseteq T,\, \K(\body^-(r)) \subseteq F \} \\ &\ \cup\{\K a\in {\KA}({\cK}) \mid {\OB}_{{\cO}, T}\models a\},\\
W_{{\cK}}(T, F)  =&\ (\, T_{{\cK}}(T, F),\, U_{{\cK}}(T, F)\,).
\end{align*}
\end{comment}
\begin{align*}
T_{{\cK}}^{(T, F)}(X, Y) =&\ \{ \K a \mid r\in {\cP},\, \K a\in \head(r),\, \body^+(r)\subseteq T\cup X,\, \K(\body^-(r)) \subseteq F\cup Y \} \\ &\ \cup\{\K a\in {\KA}({\cK}) \mid {\OB}_{{\cO}, T\cup X}\models a\},\\
U_{{\cK}}^{(T, F)}(X, Y) =&\ U_{{\cK}}(T\cup X, F\cup Y),\\
W_{{\cK}}^{(T, F)}(X, Y)  =&\ (\, T_{{\cK}}^{(T, F)}(X, Y),\, U_{{\cK}}^{(T, F)}(X, Y)\,).
\end{align*}

Note that, $W_{{\cK}}^{(T, F)}$ is monotonic, \ie, 
if $(X_1, Y_1) \sqsubseteq (X_2, Y_2)$, then 
\[
(T_{\cK}^{(T, F)}(X_1, Y_1),\, U_{\cK}^{(T, F)}(X_1, Y_1)) \sqsubseteq (T_{{\cK}}^{(T, F)}(X_2, Y_2),\, U_{\cK}^{(T, F)}(X_2, Y_2)).
\]
Notice also that each partition $(T,F)$ induces an instance operator $W_{{\cK}}^{(T, F)}$. Thus,  we have defined a family of monotonic operators. We often just write $W_{{\cK}}$, and call it a well-founded operator,
 to mean the family of  instance operators induced from partial partitions.

\begin{definition}
The {\em well-founded partition} of a normal hybrid \MKNF knowledge base ${\cK}$
 is defined by the least fixpoint of the instance operator $W_{{\cK}}^{(\emptyset, \emptyset)}$.
\end{definition}
In particular, we define
\begin{align*}
W_{{\cK}}^{(T, F)}\!\uparrow^0 &= (\emptyset, \emptyset),\\
W_{{\cK}}^{(T, F)}\!\uparrow^k &= W_{{\cK}}^{(T, F)}(W_{{\cK}}^{(T, F)}\!\uparrow^{k-1}), & k>0\\
W_{{\cK}}(T, F) &= W_{{\cK}}^{(T, F)}\!\uparrow^\infty.
\end{align*}
The well-founded partition of ${\cK}$ is $W_{\cK}(\emptyset,  \emptyset)$. If the entailment relation $ {\OB}_{{\cO},\, X}\models a$ can be computed in polynomial time, then $W_{\cK}(T, F)$ can be computed in polynomial time.

\setcounter{example}{0}
\begin{example}[Continued]
The well-founded partition of ${\cK}_1$ in Example \ref{exp:1} can be computed as follows:
\begin{align*}
T_{{\cK}_1}^{(\emptyset, \emptyset)}(\emptyset, \emptyset) &= \emptyset, & U_{{\cK}_1}^{(\emptyset, \emptyset)}(\emptyset, \emptyset) &= \{ \K c\}, &
T_{{\cK}_1}^{(\emptyset, \emptyset)}(\emptyset, \{\K c\}) &= \emptyset, & U_{{\cK}_1}^{(\emptyset, \emptyset)}(\emptyset, \{\K c\}) &= \{ \K c\}.
\end{align*}
Then $W_{\cK_1}(\emptyset, \emptyset) = W_{{\cK}_1}^{(\emptyset, \emptyset)}\!\uparrow^{\infty}\, = (\emptyset, \{\K c\})$.
\end{example}

\begin{comment}
\begin{example}
Consider ${\cK}_2 = ({\cO}_2, {\cP}_2)$, where $\pi({\cO}_2) = \neg (a\land b)$ and ${\cP}_2$ consists of
\begin{align*}
\K a &\gets \Not b. & \K b &\gets \Not a.
\end{align*}
The well-founded partition of ${\cK}_2$ can be computed as follows:
\begin{align*}
T_{{\cK}_2}(\emptyset, \emptyset) &= \emptyset, & U_{{\cK}_2}(\emptyset, \emptyset) &= \emptyset.
\end{align*}
Then, $W_{{\cK}_2}\!\uparrow^{\infty}\!(\emptyset,  \emptyset) = (\emptyset, \emptyset)$.
\end{example}
\end{comment}

\subsection{An expanding well-founded operator}

{Here, we introduce another well-founded operator
extended from $W_{\cK}$.  The idea is to apply unit propagation to increase the propagation power of $W_{\cK}$.}

Let ${\cK} = ({\cO}, {\cP})$ be a normal hybrid \MKNF knowledge base, $(T, F)$ a partial partition of ${\KA}({\cK})$. We use $\UP_{\cK}^{(T, F)}(X, Y)$ to denote the partial partition that can be derived from ${\cK}$ based on $(T\cup X, F\cup Y)$ by unit propagation. Formally, it is defined in Algorithm~\ref{alg:0}.

\begin{algorithm}[htp]\label{alg:0}
\caption{$\UP_{\cK}^{(T, F)}(X, Y)$}
append $\{\K a\in {\KA}({\cK}) \mid {\OB}_{{\cO}, T\cup X}\models a\}$ to $X$;\\
\While{there exists $r\in {\cP}$ \st {\small$\left(\left(\head(r)\cup \K(\body^-(r))\right)\setminus (F\cup Y)\right) \cup \left( \body^+(r)\setminus (T\cup X)\right) = \{\K a\}$} for some $\K a\in \KA(\cK)$}
{
  \If{$\K a \in \head(r) \cup \K(\body^-(r))$}
  {
    append $\{\K a\}$ to $X$;
  }
  \Else
  {
    append $\{\K a\}$ to $Y$;
  }
}
\If{there exists $r\in {\cP}$ \st {\small$\left(\left(\head(r)\cup \K(\body^-(r))\right)\setminus (F\cup Y)\right) \cup \left( \body^+(r)\setminus (T\cup X)\right) = \emptyset$}}
{
	\Return{$(\KA(\cK), \KA(\cK))$};
}
\Return{$(X, Y)$};
\end{algorithm}%

Then we introduce the expanding well-founded operator $E_{\cK}^{(T, F)}$ of ${\cK}$ as follows:
\begin{align*}
E_{{\cK}}^{(T, F)}(X, Y) = \UP_{\cP}^{(T, F)}(X, Y) \sqcup (\,\emptyset,\,\,  U_{{\cK}}^{(T, F)}(X, Y)\,).
\end{align*}

For example, 
consider a hybrid \MKNF knowledge base $\cK = (\cO,\cP)$, where $\pi(\cO) = \top$ and $\cP = \{\K p \leftarrow \bfnot \, q\}$.  
By the definition of $E_\cK^{(\emptyset, \{\K p\})}$, we have
$\UP^{(\emptyset, \{\K p\})}(\emptyset, \emptyset) = (\{\K q\}, \emptyset)$ and
$(\emptyset, U_{{\cK}}^{(\emptyset, \{\K p\})}(\emptyset, \emptyset)) = (\emptyset, \{\K q\})$,
and thus $E_{{\cK}}^{(\emptyset, \{\K p\})}(\emptyset, \emptyset)$ is inconsistent.  For this example, the result shows that no \MKNF model $M$ exists under the condition that $M\models_\MKNF \neg \K p$.

Since $\UP_{\cP}^{(T, F)}$ is monotonic, $E_{{\cK}}^{(T,F)}$ is monotonic as well. Again, above we have  defined a family of monotonic operators. We may write $E_{{\cK}}$, and call it a well-founded operator,
to mean the family of these instance operators.

\begin{definition}
The {\em expanding well-founded partition} of a normal hybrid \MKNF knowledge base ${\cK}$ is defined by the least fixpoint of the instance operator $E_{\cK}^{(\emptyset, \emptyset)}$.
\end{definition}
Similarly, we define
\begin{align*}
E_{{\cK}}^{(T, F)}\!\uparrow^0 &= (\emptyset, \emptyset),\\
E_{{\cK}}^{(T, F)}\!\uparrow^k &= E_{{\cK}}^{(T, F)}(E_{{\cK}}^{(T, F)}\!\uparrow^{k-1}), & k>0\\
E_{\cK}(T, F) &= E_{{\cK}}^{(T, F)}\!\uparrow^{\infty}.
\end{align*}
The expanding well-founded partition of ${\cK}$ is $E_{{\cK}}(\emptyset,  \emptyset)$. If the entailment relation $ {\OB}_{{\cO},\, X}\models a$ can be computed in polynomial time, then $E_{\cK}(T, F)$ can be computed in polynomial time.

Note that, since $T_{\cK}^{(T, F)}(X, Y) \sqsubseteq \UP_{\cK}^{(T, F)}(X, Y)$, the expanding well-founded operator $E_\cK$ is an extension of the well-founded operator $W_\cK$.
\begin{proposition}\label{prop:new1}
Let ${\cK}$ be a normal hybrid \MKNF knowledge base and $(T, F)$ a partial partition of $\KA(\cK)$.
$W_{\cK}^{(T, F)}(X, Y) \sqsubseteq E_{\cK}^{(T, F)}(X, Y)$ and $W_{\cK}(T, F) \sqsubseteq E_{\cK}(T, F)$.
\end{proposition}

The following example shows that $W_{\cK}(T, F) \sqsubset E_{\cK}(T, F)$ is possible.
\setcounter{example}{0}
\begin{example}[Continued]
The expanding well-founded partition of ${\cK}_1$ can be computed as follows:
\begin{align*}
\UP_{{\cP}_1}^{(\emptyset, \emptyset)}(\emptyset, \emptyset) &= (\emptyset, \emptyset), & U_{{\cK}_1}^{(\emptyset, \emptyset)}(\emptyset, \emptyset) &= \{ \K c\},\\
\UP_{{\cK}_1}^{(\emptyset, \emptyset)}(\emptyset, \{\K c\}) &= (\{\K b\}, \{\K a, \K c\}),
& U_{{\cK}_1}^{(\emptyset, \emptyset)}(\emptyset, \{\K c\}) &= \{ \K c\}.
\end{align*}
Then $E_{\cK_1}(\emptyset, \emptyset) =  (\{\K b\}, \{\K a, \K c\})$, which corresponds to the unique \MKNF model of~${\cK}_1$.
\end{example}

\subsection{Relations to coherent well-founded partition}

In this subsection, we show the relations of the new well-founded operators proposed in this paper with the one based on the alternating fixpoint construction.

\begin{theorem}\label{them:eq}
Let ${\cK}$ be a normal hybrid \MKNF knowledge base.
{$W_{{\cK}}(\emptyset,  \emptyset)=({\boldP}_\omega,\, {\KA}({\cK})\setminus {\boldN}_\omega)$.}
\end{theorem}

From Proposition~\ref{prop:new1}, we have $({\boldP}_\omega,\, {\KA}({\cK})\setminus {\boldN}_\omega)\sqsubseteq E_{{\cK}}(\emptyset,  \emptyset)$.

The above theorem shows that the well-founded partition is equivalent to the coherent well-founded partition.
On the other hand, given a partial partition $(T, F)$ for a normal hybrid \MKNF knowledge base ${\cK}$, $W_{{\cK}}(T, F)$ returns an expansion of $(T, F)$. Similarly, for the purpose of adopting alternating fixpoint construction for constraint propagation, 
we may attempt to define ${\boldP}^{(T, F)}_\omega$ and ${\boldN}^{(T, F)}_\omega$ from the sequences ${\boldP}_i$ and ${\boldN}_i$ with ${\boldP}_0 = T$ and ${\boldN}_0 = {\KA}({\cK}) \setminus F$.

{As shown below,}  $W_{{\cK}}(T, F)$ may not coincide with $({\boldP}^{(T, F)}_\omega, {\KA}({\cK})\setminus {\boldN}^{(T, F)}_\omega)$.

\setcounter{example}{1}
\begin{example}
\label{2}
{Consider ${\cK}_2= ({\cO}_2, {\cP}_2)$, where $\pi
({\cO}_2) = (a\imply b)$ and ${\cP}_2$ consists of }
$$
\begin{array}{ll}
\K a \gets \Not c. ~~~~~~~\K c \gets \Not a. ~~~~~~~\K b \gets \K b.
\end{array}
$$
We have 
$W_{{\cK}_2}(\emptyset, \{\K b\}) = (\{\K c\}, \{\K a,\, \K b\})$, while 
\begin{align*}
{\boldP}^{(\emptyset, \{\K b\})}_0 &= \emptyset,  &    {\boldN}^{(\emptyset, \{\K b\})}_0 &= \{ \K a,\K c\}, &
{\boldP}^{(\emptyset, \{\K b\})}_1 &= \emptyset,  &     {\boldN}^{(\emptyset, \{\K b\})}_1 &= \{ \K a, \K b, \K c\}, & &\cdots
\end{align*}
Therefore,
$({\boldP}^{(\emptyset, \{\K b\})}_\omega, {\KA}({\cK}_2)\setminus {\boldN}^{(\emptyset, \{\K b\})}_\omega)= (\emptyset, \{\K b\})$. 
\end{example}

The next example shows that, when applied to an arbitrary partial partition, the alternating fixpoint construction may not converge. 

\begin{example}
Consider ${\cK}_3 = ({\cO}_3, {\cP}_3)$, where $\pi({\cO}_3) = (a\imply b)$ and ${\cP}_3$ consists of
$$
\begin{array}{ll}
\K a \gets \Not c. ~~~~~~~ \K c \gets \Not a. ~~~~~~~ \K a \gets \Not b.
\end{array}
$$
Let $(T,F) = (\emptyset, \{\K b\})$.  Then
\begin{multline*}
W_{{\cK}_3}(T,F)
= 
W_{{\cK}_3}^{(T, F)}\!\uparrow^{2}\, = 
W_{{\cK}_3}^{(T, F)} (\{\K a\}, \{\K a, \K b, \K c \}) 
 =(\{\K a,\K b,\K c\}, \{\K a, \K b, \K c\}).
\end{multline*}
However, the sequences ${\boldP}^{(\emptyset, \{\K b\})}_i$ and ${\boldN}^{(\emptyset, \{\K b\})}_i$ do not converge.
\begin{align*}
{\boldP}^{(\emptyset, \{\K b\})}_0 &= \emptyset, & {\boldN}^{(\emptyset, \{\K b\})}_0 &= \{\K a,\, \K c\}, &
{\boldP}^{(\emptyset, \{\K b\})}_1 &= \{ \K a,\, \K b\}, & {\boldN}^{(\emptyset, \{\K b\})}_1 &= \{\K a,\, \K b,\, \K c\},\\
{\boldP}^{(\emptyset, \{\K b\})}_2 &= \emptyset, & {\boldN}^{(\emptyset, \{\K b\})}_2 &= \{ \K a, \K b\}, &
{\boldP}^{(\emptyset, \{\K b\})}_3 &= \{ \K a,\, \K b\}, & {\boldN}^{(\emptyset, \{\K b\})}_3 &= \{\K a,\, \K b,\, \K c\},\\
&\cdots
\end{align*}
Note that $({\boldP}^{(\emptyset, \{\K b\})}_\omega,\, {\KA}({\cK}_3)\setminus {\boldN}^{(\emptyset, \{\K b\})}_\omega)= (\{\K a,\, \K b\},$ $\{\K b, \K c\})$. 
\end{example}

Note that the non-converging issue does not arise when the alternating fixpoint construction commences only from the least partition $(\emptyset, \emptyset)$. 
However, for the goal of constraint propagation, 
converging must be guaranteed when applied to arbitrary partitions. 

\begin{comment}
\begin{example}
Consider ${\cK}_5 = ({\cO}_5, {\cP}_5)$, where $\pi({\cO}_5) = \top$ and ${\cP}_5$ consists of
\begin{align*}
\K a &\gets \Not c. & \K c&\gets \Not a. 
\end{align*}
$W_{{\cK}_5}\!\uparrow^{\infty}\!(\emptyset, \{\K c\}) = (\{\K a\},\, \{\K c\})$. However, the sequences ${\boldP}^{(\emptyset, \{\K c\})}_i$ and ${\boldN}^{(\emptyset, \{\K c\})}_i$ do not converge.
\begin{align*}
{\boldP}^{(\emptyset, \{\K c\})}_0 &= \emptyset, & {\boldN}^{(\emptyset, \{\K c\})}_0 &= \{\K a\},\\
{\boldP}^{(\emptyset, \{\K c\})}_1 &= \{ \K a\}, & {\boldN}^{(\emptyset, \{\K c\})}_1 &= \{\K a,\, \K c\},\\
{\boldP}^{(\emptyset, \{\K c\})}_2 &= \emptyset, & {\boldN}^{(\emptyset, \{\K c\})}_2 &= \{ \K a\}, \\
{\boldP}^{(\emptyset, \{\K c\})}_3 &= \{ \K a\}, & {\boldN}^{(\emptyset, \{\K c\})}_3 &= \{\K a,\, \K c\},\\
{\boldP}^{(\emptyset, \{\K c\})}_2 &= \emptyset, & {\boldN}^{(\emptyset, \{\K c\})}_3 &= \{\K a\},\\
& \cdots
\end{align*}
We can still get $({\boldP}^{(\emptyset, \{\K c\})}_\omega,\, {\KA}({\cK})\setminus {\boldN}^{(\emptyset, \{\K c\})}_\omega)= (\{\K a\}, \{\K c\})$.
On the other hand, $W_{{\cK}_5}\!\uparrow^{\infty}\!(\emptyset, \emptyset) = ({\boldP}_\omega,\, {\KA}({\cK}_5)\setminus {\boldN}_\omega) = (\emptyset, \emptyset)$.
\end{example}
\end{comment}

\begin{theorem}
\label{3.3}
Let ${\cK}$ be a normal hybrid \MKNF knowledge base and $(T, F)$ a partial partition of ${\KA}({\cK})$.
$({\boldP}^{(T, F)}_i,\, {\KA}({\cK})\setminus {\boldN}^{(T, F)}_i) \sqsubseteq W_{{\cK}}(T,  F) \sqsubseteq E_{{\cK}}(T,  F)$, for each $i>0$.
\end{theorem}

\section{Computing \MKNF Models}

We show that both well-founded operators can be used to compute \MKNF models of a normal hybrid \MKNF knowledge base in a DPLL-based procedure. We first provide some properties.

\begin{theorem}\label{them:compute}
Let ${\cK}$ be a normal hybrid \MKNF knowledge base, $(T, F)$ a partial partition of ${\KA}({\cK})$, $W_{{\cK}}(T, F) = (T^*_W, F^*_W)$, and $E_{{\cK}}(T, F) = (T^*_E, F^*_E)$. Then for any \MKNF model $M$ of ${\cK}$ with $M\models_{\MKNF} \bigwedge_{\K a\in T} \K a \land \bigwedge_{\K b\in F} \neg \K b$, 
\begin{itemize}
\item $M\models_\MKNF \bigwedge_{\K a\in T^*_W} \K a \land \bigwedge_{\K b\in F^*_W} \neg \K b$, and 
\item $M\models_\MKNF \bigwedge_{\K a\in T^*_E} \K a \land \bigwedge_{\K b\in F^*_E} \neg \K b$.
\end{itemize}
\end{theorem}

The theorem can be proved from Proposition~\ref{prop:2}, \ie, given an unfounded set $U$ of $\cK$ \wrt $(T, F)$, if $M$ is an \MKNF model of ${\cK}$ satisfying $(T, F)$, then $M \models_\MKNF \neg K b$ for each $b\in U$.
\begin{comment}
\begin{proof}
The theorem can be proved from the fact that $M\models_\MKNF \neg \K b$ for each $b\in U$ where $U$ is an unfounded set of ${\cK}$ \wrt $(T, F)$.
\end{proof}
\end{comment}

\begin{corollary}
Let ${\cK}= ({\cO}, {\cP})$ be a normal hybrid \MKNF knowledge base, $(T_W, F_W)$ the well-founded partition of $\cK$, and $(T_E, F_E)$ the expanding well-founded partition of $\cK$. 
\begin{itemize}
\item If $T_W\cup F_W = {\KA}({\cK})$ and $T_W\cap F_W = \emptyset$, then $M=\{ I\mid I\models  {\OB}_{{\cO}, T_W}\}$ is the only \MKNF model of ${\cK}$.
\item If $T_E\cup F_E = {\KA}({\cK})$ and $T_E\cap F_E = \emptyset$, then $M=\{ I\mid I\models  {\OB}_{{\cO}, T_E}\}$ is the only \MKNF model of ${\cK}$.
\item If $T_W\cap F_W \neq\emptyset$ or $T_E\cap F_E\neq\emptyset$, then ${\cK}$ does not have an \MKNF model.
\end{itemize}
\end{corollary}

Algorithm~\ref{alg:1} gives a DPLL-based procedure to compute an \MKNF model of a normal hybrid \MKNF knowledge base~$\cK$ by a call over partition~$(\emptyset,\emptyset)$, if one exists, and returns $\tt false$ otherwise, where $\WFM_{\cK}(T, F)$ is either $W_{{\cK}}(T, F)$ or $E_{{\cK}}(T, F)$. By backtracking, the algorithm can be extended to compute all \MKNF models of~${\cK}$.

\begin{algorithm}[htp]\label{alg:1}
\caption{$\text{\it solver}({\cK}, (T, F))$}
$(T, F) := \WFM_{\cK}(T, F) \sqcup (T, F)$;\\
\If{$T\cap F\neq\emptyset$}
{
  \Return{$\text{\tt false}$};
}
\ElseIf{$T\cup F={\KA}({\cK})$}
{
  \Return{$\text{\tt true}$};
}
\Else
{
  choose a $\bfK$-atom $\K a$ from ${\KA}({\cK})\setminus (T\cup F)$;\\
  \If{$\text{\it solver}({\cK}, (T\cup \{\K a\}, F))$}
  {
    \Return{$\text{\tt true}$};
  }
  \Else
  {
    \Return{$\text{\it solver}({\cK}, (T, F\cup \{\K a\}))$};
  }
}
\end{algorithm}%

\begin{theorem}
Let ${\cK}= ({\cO}, {\cP})$ be a normal hybrid \MKNF knowledge base.
If $\text{\it solver}({\cK}, (\emptyset, \emptyset))$ returns $\text{\tt true}$ and $(T, F)$ is the corresponding result in Algorithm~\ref{alg:1}, then  $M=\{ I\mid I\models  {\OB}_{{\cO}, T}\}$ is an \MKNF model of ${\cK}$.
If $\text{\it solver}({\cK}, (\emptyset, \emptyset))$ returns $\text{\tt false}$, then ${\cK}$ does not have an \MKNF model.
\end{theorem}

\section{Simplifying Hybrid \MKNF Knowledge Bases}

The well-founded model of a logic program~\cite{van1991well} can be used to simplify the program so that the resulting program would no longer contain atoms appearing in the model. The well-founded model has been used in grounding engines of most ASP solvers to simplify programs~\cite{baral2003knowledge}.
{In general, however,} we cannot extend the well-founded model by a consequence, \ie, a set of literals that are satisfied by every answer set, in these grounding engines, as a consequence may not be used to simplify the given program~\cite{ji2015simplifying}. 

Here we show that the well-founded partition, $W_{{\cK}}(\emptyset, \emptyset)$, can be used to simplify the rule base of the normal hybrid \MKNF knowledge base $\cK$, while it is not safe to do so for the expanding well-founded partition $E_{{\cK}}(\emptyset, \emptyset)$. Thus, we should simplify the rule base of $\cK$ by $W_{{\cK}}(\emptyset, \emptyset)$, before we apply Algorithm~\ref{alg:1} to compute MKNF models of~$\cK$, in which the stronger operator $E_\cK$ should be used as the constraint propagator.

We first introduce a method to simplify a rule base by a partial partition.
Let ${\cK} = ({\cO}, {\cP})$ be a normal hybrid \MKNF knowledge base and $(T, F)$ a partial partition of ${\KA}({\cK})$. We denote ${\cK}^{(T, F)} = ({\cO}^{(T, F)}, {\cP}^{(T, F)})$ to be the hybrid \MKNF knowledge base reduced from ${\cK}$ under $(T, F)$, where ${\cO}^{(T, F)} = {\cO} \cup \{ a \mid \K a\in T\}$ and ${\cP}^{(T, F)}$ is obtained from ${\cP}$ by deleting:
\begin{enumerate}
  \item each \MKNF rule $r$ that satisfies one of the following conditions:
  \begin{itemize}
    \item $\body^+(r)\cap F\neq\emptyset$,
    \item $\bfK(\body^-(r)) \cap T \neq\emptyset$, or
    \item $\head(r)\cap T\neq\emptyset$,
  \end{itemize}
  \item all formulas of the form $\K a$ in the heads of the remaining rules with $\K a \in F$,
  \item all formulas of the form $\K a$ in the bodies of the remaining rules with $\K a\in T$,
  \item all formulas of the form $\Not a$ in the bodies of the remaining rules with $\K a\in F$.
\end{enumerate}
Notices that ${\cP}^{(T, F)}$ contains no first-order atom $a$ with $\K a\in T\cup F$.

\begin{comment}
In general, however, an \MKNF model of ${\cK}^{(T, F)}$ may not be an \MKNF model of ${\cK}$.
\begin{example}
Consider ${\cK}_4= ({\cO}_4, {\cP}_4)$, where $\pi({\cO}_4) = \top$ and ${\cP}_4$ consists of
\begin{align*}
\K c &\gets \K a. & \K a &\gets \Not b. & \K b &\gets \Not a.
\end{align*}
The reduction ${\cK}_4^{(\{\K c\}, \emptyset)}$ is $( \{ c\},\, \{ \K a \gets \Not b. \ \K b\gets \Not a.\})$. Note that, $M = \{ I \mid I\models c \land b\}$ is an \MKNF model of ${\cK}_4^{(\{\K c\}, \emptyset)}$. However, $M$ is not an \MKNF model of~${\cK}_4$.
\end{example}
\end{comment}

The following theorem shows that, if $(T, F)$ is the well-founded partition, then ${\cK}$ and ${\cK}^{(T, F)}$ have the same set of \MKNF models.

\begin{theorem}
\label{simplify}
Let ${\cK}$ be a normal hybrid \MKNF knowledge base and $(T, F)$ the well-founded partition of ${\cK}$. An \MKNF interpretation $M$ is an \MKNF model of ${\cK}$ if{f} $M$ is an \MKNF model of ${\cK}^{(T, F)}$.
\end{theorem}

The following example shows that, it is not safe to use the expanding well-founded partition, $E_{{\cK}}(\emptyset, \emptyset)$,  to simplify the rule base of a normal hybrid \MKNF knowledge base~$\cK$.

\begin{example}
Consider ${\cK}_4 = (\cO_4, \cP_4)$, where $\pi(\cO_1) = \neg c$ and $\cP_4$ consists of
$$
\begin{array}{ll}
\K a \gets \K d.  ~~~~~~~\K b \gets \Not d.  ~~~~~~~\K d \gets \Not b.  ~~~~~~~\K c \gets \Not a.
\end{array}
$$
Similar to Example~\ref{exp:1}, $E_{{\cK_4}}(\emptyset, \emptyset) = (\{\K a\}, \{\K c\})$. Then
$\cO_4^{(\{\K a\}, \{\K c\})} = \cO_4 \cup \{a\}$ and $\cP_4^{(\{\K a\}, \{\K c\})} = \{\,\K b \gets \Not d. \ \K d \gets \Not b. \,\}$. 
It is easy to verify that $M = \{ I\mid I\models a\land d \land \neg c\}$ is the only \MKNF model of $\cK_4$. However, $M' = \{I\mid I\models a \land b \land \neg c\}$ is also an \MKNF model of $\cK_4^{E_{{\cK_4}}(\emptyset, \emptyset)}$.
\end{example}

\section{Related Work and Discussion}

\noindent
{\bf Well-founded operators and three-valued semantics}

For a normal logic program, the well-founded model uniquely exists, which can be computed by the operator based on alternating fixpoint construction \cite{DBLP:journals/jcss/Gelder93}
as well as by the one based on unfounded sets \cite{van1991well}.  However, 
for normal hybrid \MKNF knowledge bases different well-founded operators are possible. In particular, we have shown that the well-founded operator $E_\cK$ proposed in this paper is  stronger than either the operator $W_\cK$ or the operator based on the alternating fixpoint construction. It is interesting to note that $E_\cK$ sometimes generates an \MKNF model directly, whereas a weaker operator
 computes the well-founded partition that is not even a three-valued \MKNF model (as defined in \cite{knorr2011local}).

\begin{example}
 \label{6}
Consider $\cK = (\cO, \cP)$, where $\pi (\cO) = (unemployed \supset \neg employed) \wedge unemployed$ and $\cP$ is
$$
\begin{array}{ll}
\K employed \leftarrow \K salary.    ~~~~~~~~~~~~~~~~~~~\K volunteer \leftarrow \K work, \, \bfnot\, salary.\\
\K salary \leftarrow \K work, \, \bfnot \, volunteer. ~~~~\K work \leftarrow.
\end{array}
$$
While the expanding well-founded partition 
assigns $\K employed, \K salary$ to {\tt false}, and $\K volunteer$ and $\K work$ to {\tt true}, which corresponds to an \MKNF model, the well-founded partition generated by $W^{(\emptyset,\emptyset)}_\cK$, as well as the coherent well-founded partition {generated by the alternating fixpoint construction,} assigns $\K employed$ to {\tt false}, $\K work$ to {\tt true}, and the rest to {\tt undefined}, which does not correspond to a three-valued \MKNF model. Intuitively, the reason is that the first rule is not satisfied in three-valued logic, as its head is false and its body is undefined. An interesting observation is that a partial \MKNF interpretation that can be used to simplify a hybrid \MKNF knowledge base need not be a three-valued \MKNF model. 
\end{example}

In general,  a normal hybrid \MKNF knowledge base may not possess a three-valued \MKNF model.  As a further complication,  though the well-founded model of a normal logic program $P$ equals the intersection of all three-valued models of $P$, {it can be shown} that even the problem of 
determining the existence of a three-valued \MKNF model for a normal hybrid \MKNF knowledge base is NP-complete (assuming that the underlying DL is trackable). All these indicate that the notion of well-founded operators  for hybrid \MKNF knowledge bases is in general a non-trivial research issue. 

\vspace{.1in}
\noindent
{\bf Relation to other approaches to combining DLs with ASP}

In \cite{motik2010reconciling}, the authors extensively discussed how some of the popular approaches to combining DLs and ASP can be captured by hybrid \MKNF knowledge bases. Answer set programs with external sources  is one approach that provides some recent implementation techniques \cite{DBLP:conf/ijcai/EiterKRW16}. Though closely related, their techniques do not directly apply to computing \MKNF models, since in general hybrid \MKNF knowledge bases represent a tighter integration. For example,  in answer set programs with external sources DL predicates cannot appear in the heads of rules, which is the case in Example \ref{6}. 

A further question of interest is whether the approximation fixpoint theory (AFT) of \cite{DeneckerMT04}) can be applied to define well-founded operators as proposed in this paper.  How to apply AFT to hybrid MKNF knowledge bases {is a nontrivial research issue. One of the difficulties is that} in the current formalism these operators can only be mappings on consistent elements in a bilattice 
(or, they can be ``symmetric" operators; also see \cite{RR14}). The work on FO(ID) \cite{VennekensDB10}
 is a very different but loose combination, where 
 the rule component is used to define concepts, whereas the FO component asserts additional properties of the defined concepts.  All formulas in FO(ID) are interpreted under closed world assumption. Thus, hybrid \MKNF knowledge bases and FO(ID) have some fundamental differences in basic ideas.

\section{Conclusion and Future Work}
{The goal of this paper is to address the critical issue of constraint propagation in a DPLL-style search engine for reasoning with hybrid \MKNF knowledge bases.
We first proposed the notion of unfounded sets for normal hybrid \MKNF knowledge bases, based on which we introduced two well-founded operators with different powers of propagation. The first well-founded operator computes the greatest unfounded set \wrt a partial partition to generate the false $\bfK$-atoms and uses rules to generate true $\bfK$-atoms. The second one in addition applies unit propagation to infer more truth values.}
We showed that both operators compute more truth values than \citeANP{knorr2011local}'s operator when applied to arbitrary partitions, and this is achieved 
without increasing the computational data complexity.
We then defined a DPLL search engine to compute  \MKNF models by employing either of the new operators as a propagator. We also contrasted the two operators, one of which can be used to simplify
the given hybrid \MKNF knowledge base and the other, as a stronger propagator, is best used as a propagator in a DPLL search engine. 

Our next step is to extend the well-founded operators to disjunctive hybrid \MKNF knowledge bases. We are also interested in how to incorporate conflict-directed backtracking and clause learning into such a DPLL engine, and we are considering to implement and experiment with a solver based on the discoveries.

\bibliographystyle{acmtrans}
\bibliography{ref}


\newpage

\appendix
\section{Unfounded sets by Knorr et al.}
\label{unfounded-compare}

In the proof of Proposition 7 of \cite{knorr2011local}, conditions are given which are similar to, but do not coincide with, the conditions in Def \ref{unfounded} of this paper. In fact, as shown below, when applied to arbitrary partitions, their definition becomes problematic for our purpose. 

Let $P_n$, $N_n$ be the  sequences of $P_\omega$ and $N_\omega$, \ie, the  sequences in computing the coherent well-founded partition. Let $U$ be the set of all ${\K H} \not \in \Gamma'_{{\cK}}(\boldP_n)$. Note that ${\OB}_{\cO,P_n}$ must be consistent. Then, for each ${\K H} \in U$, the following conditions are satisfied:

\begin{itemize}
\item[({U1})] for each ${\K H} \leftarrow body$ in $\cP$, at least one of the folllowing holds:
 \begin{itemize}
	\item[({U1a})] some modal $\K$-atom ${\K A}$ appears in $body$ and in $U \cup  {\KA}({\cK}) \setminus \boldN_n$;
	\item[({U1b})] some modal $\Not$-atom $\Not B$ appear in $body$ and in $\boldP_n$;
          \item[({U1c})] ${\OB}_{\cO,P_n} \models \neg H$.
\end{itemize}
\item[({U2})] for each $S$ with $S \subseteq \boldP_n$, on which $\K H$ depends, there is at least one modal $\K$-atom $\K A$ such that ${\OB}_{\cO,S\setminus {\K A}} \not \models H$ and ${\K A} \in U \cup  {\KA}({\cK}) \setminus \boldN_n$.
\end{itemize}

In a footnote, the authors commented that these conditions resemble the notion of unfounded sets in \cite{van1991well}.

By this definition, let us consider Example \ref{2} again. 

\begin{example}
Recall ${\cK}_2= ({\cO}_2, {\cP}_2)$, where $\pi({\cO}_2) = (a\imply b)$ and ${\cP}_2$ consists of
$$
\begin{array}{ll} 
\K a \gets \Not c.  ~~~~~~~~~\K c \gets \Not a. ~~~~~~~~~\K b \gets \K b.
\end{array}
$$
By the alternating fixpoint construction, its coherent well-founded partition is $(\emptyset, \{\K a,\K b,\K c\})$, \ie, it has  all $\bfK$-atoms undefined, which is correctly captured by the alternating fixpoint construction as well as by their definition of unfounded set.  Thus, their notion of unfounded set serves the purpose of proving the properties of a well-founded semantics. 

However, the difference shows up when applied to arbitrary partitions.  Let 
$(T,F)= (\emptyset,\{\K b\})$. Then, based on the above definition, the unfounded set is $\emptyset$. That is, even that $\K b$ is false in the given partition is lost in the result of computing unfounded set. In contrast, by our definition, Definition \ref{unfounded}, the unfounded set is $\{\K a, \K b\}$. 

\end{example}

\section{Proofs}
\label{proof-appendix}
{\bf Proposition \ref{union}}\\
Let ${\cK}$ be a normal hybrid \MKNF knowledge base, $(T, F)$ a partial partition of ${\KA}({\cK})$. If $X_1$ and $X_2$ are unfounded sets of ${\cK}$ \wrt $(T, F)$, then $X_1\cup X_2$ is an unfounded set of ${\cK}$ \wrt $(T, F)$.
\begin{proof}
For each ${\K a} \in X_1$ and the corresponding \MKNF rule $r$, that $\body^+(r)\cap X_1 \neq \emptyset$ implies $\body^+(r)\cap (X_1\cup X_2) \neq\emptyset$. Similarly for each $\K a\in X_2$, then $X_1\cup X_2$ is also an unfounded set of ${\cK}$ \wrt~$(T, F)$.
\end{proof}

\vspace{.1in}
\noindent
{\bf Proposition \ref{prop:2}}\\
Let ${\cK}$ be a normal hybrid \MKNF knowledge base, $(T, F)$ a partial partition of ${\KA}({\cK})$, and $U$ an unfounded set of ${\cK}$ \wrt $(T, F)$. For any \MKNF model $M$ of ${\cK}$ with $M\models_\MKNF \bigwedge_{\K a\in T} \K a \land \bigwedge_{\K b\in F} \neg \K b$, $M\models_\MKNF \neg \K u$ for each $\K u\in U$.
\begin{proof}
Assume that there exists such an \MKNF model $M$ with $M\models_\MKNF \K u$ for some $\K u\in U$. Let $U^*$ be the greatest unfounded set of ${\cK}$ \wrt $(T, F)$ and  
\[
M' =
\{ I' \mid I'\models  {\OB}_{{\cO}, T} \text{ and } I'\models a, \forall a\in {\KA}({\cK})\setminus U^* \text{ with $M\models_\MKNF \K a$}\}.
\]
Note that, $ {\OB}_{{\cO}, T}\not\models u$ for each $u\in U^*$, and thus $M' \supset M$. 

Clearly, $(I', M', M)\models {\K \pi({\cO})}$ for each $I'\in M'$, $M'\models_\MKNF \neg {\K u}$ for each $u\in U^*$, and 
$\{ {\K a} \in {\KA}({\cK}) \mid M\models_\MKNF {\K a}\} \setminus U^* =  \{{\K a} \in {\KA}({\cK}) \mid M'\models_\MKNF {\K a} \}$.  Let us denote the last set by~$T^*$.

For each $r\in {\cP}$, if $\body^+(r)\subseteq T^*$ and $\K(\body^-(r))\cap$ $ T^*= \emptyset$, then $\head(r) \subseteq T^*$ and $\head(r) \cap U^*=\emptyset$. So $M'\models_\MKNF \pi(r)$.
It then follows that   $(I', M', M)\models \pi({\cK})$ for each $I'\in M'$, which contradicts the precondition that $M$ is an \MKNF model of ${\cK}$. Therefore, $M\models_\MKNF \neg \K u$ for each $\K u\in U$.
\end{proof}

\vspace{.1in}
\noindent
{\bf Proposition \ref{unfounded-free}}\\
Let ${\cK} = ({\cO}, {\cP})$ be a normal hybrid knowledge base and $M$ an \MKNF model of $\cK$. Define $(T,F)$ by $T = \{\K a \in {\KA}({\cK}) \mid M \models_\MKNF \K a\}$ and  $F = {\KA}({\cK})\setminus T$.  Then, $F$ is the greatest unfounded set of 
${\cK}$  \wrt $(T,F)$.
\begin{proof}
Let $U^*$ be the greatest unfounded set of 
${\cK}$  \wrt $(T,F)$. We prove $F = U^*$.  That $U^* \subseteq F$ follows from 
Proposition \ref{prop:2} under the special case that the given partition $(T,F)$ satisfies 
$T = \{\K  a \in {\KA}({\cK}) \mid M \models_\MKNF \K a\}$ and  $F = {\KA}({\cK})\setminus T$.

To show $F \subseteq U^*$, assume $\K a \not \in U^*$, from which for any unfounded set $U$ of ${\cK}$  \wrt $(T,F)$,
$\K a \not \in U$. By definition (Def.~\ref {unfounded}), for each $R \subseteq {\cP}$ such that $head(R) \cup {\OB}_{{\cO}, T} \models \K a$ and $head(R) \cup {\OB}_{{\cO}, T} \cup \{\neg b\}$ is consistent for any $\K b \in F$, no rule $r \in R$ satisfies any of the 
three conditions in Def.~\ref{unfounded}, which implies $body^+(r) \subseteq T$ and $\K (body^-(r)) \subseteq F$ and, as $M$ is an \MKNF model of $\cK$, $head(R) \subseteq T$ and it follows
 $\K a \in T$. By definition,  that $\K a \in T$ implies $\K a \not \in F$. 
\end{proof}

\vspace{.1in}
\noindent
{\bf Theorem \ref{theorem1}}

\noindent
Let ${\cK}$ be a normal hybrid \MKNF knowledge base and $(T, F)$ a partial partition of ${\KA}({\cK})$. $U_{{\cK}}(T, F) = {\KA}({\cK}) \setminus {\atmost}_{{\cK}}(T, F)$.

\begin{proof}
We first prove that ${\KA}({\cK}) \setminus {\atmost}_{{\cK}}(T, F)$ is an unfounded set of ${\cK}$ \wrt $(T, F)$, then we prove that for any other unfounded set $U$, $U \subseteq {\KA}({\cK}) \setminus {\atmost}_{{\cK}}(T, F)$.

(1) Let $X = {\KA}({\cK}) \setminus {\atmost}_{{\cK}}(T, F)$. If $X$ is not an unfounded set of ${\cK}$ \wrt $(T, F)$, then there exist a {$\bfK$-atom} $\K a\in X$ and a set of \MKNF rules $R\subseteq {\cP}$ such that $\head(R)\cup  {\OB}_{{\cO}, T} \models a$ and $\head(R)\cup  {\OB}_{{\cO}, T} \cup \{\neg b\}$ is consistent for each $\K b\in F$, and for each $r\in R$:
\begin{description}
  \item[$\ \ $]
\begin{itemize}
  \item $\body^+(r) \cap F= \emptyset$,
  \item $\K(\body^-(r))\cap T=\emptyset$, and
  \item $\body^+(r)\cap X=\emptyset$.
\end{itemize}
\end{description}
Note that for each $r\in R$, $\body^+(r) \subseteq \atmost_{{\cK}}(T, F)$.
Let $Y = \{ {\K h} \mid h\in \head(R)\}$. 
From the definition of $V_{{\cK}}^{(T, F)}$, $Y \subseteq \atmost_{{\cK}}(T, F)$. It follows ${\K a} \in \atmost_{{\cK}}(T, F)$, which contradicts the precondition that ${\K a} \in {\KA}({\cK}) \setminus {\atmost}_{{\cK}}(T, F)$. So $X$ is an unfounded set of ${\cK}$ \wrt $(T, F)$.

(2) For the sake of contradiction, assume $U$ is an unfounded set of ${\cK}$ \wrt $(T, F)$ such that $U\not\subseteq {\KA}({\cK}) \setminus \atmost_{{\cK}}(T, F)$. Then there exists a $\bfK$-atom ${\K a} \in U$ such that 
${\K a} \in {\atmost}_{{\cK}}(T, F)$.

(a) If there exists an \MKNF rule $r\in {\cP}$, ${\K a}\in \head(r)$, $\body^+(r)\subseteq \atmost_{{\cK}}(T, F)$, $\body^+(r)\cap F=\emptyset$, $\K(\body^-(r))\cap T=\emptyset$, and $\{ a, \neg b\}\cup  {\OB}_{{\cO}, T}$ is consistent for each $\K b \in F$, then $\body^+(r)\cap U\neq\emptyset$. 

If $\{ {\K a}\} = \body^+(r)\cap U$, then there exists another \MKNF rule $r'\in {\cP}$ with  ${\K a} \in \head(r')$, $\body^+(r')\subseteq \atmost_{{\cK}}(T, F)$, $\body^+(r')\cap F=\emptyset$, $\K(\body^-(r'))\cap T=\emptyset$, and $\{ a, \neg b\}\cup  {\OB}_{{\cO}, T}$ is consistent for each $\K b \in F$. The process can continue until there exists such an \MKNF rule $r^*$ with $\{\K a\} \neq \body^+(r^*)\cap U$. 

If $\{\K a\} \neq\body^+(r)\cap U$, then there exists another $\bfK$-atom $\K a_1 \in U\cap atmost_{{\cK}}(T, F)$. The argument can repeat indefinitely, which results in a contradiction  to the precondition that the set ${\KA}({\cK})$ is finite. So there does not exist such an \MKNF rule and Case (a) is impossible.

(b) If $ {\OB}_{{\cO}, \atmost_{{\cK}}(T, F)}\models a$, then for each set of \MKNF rules $R\subseteq {\cP}$ with $\{\K h\mid h\in \head(R)\} \subseteq \atmost_{{\cK}}(T, F)$, $ {\OB}_{{\cO}, \{\K h\mid h\in\head(R)\}}\models a$, and for each $r\in R$, $\body^+(r)\subseteq \atmost_{{\cK}}(T, F)$, $\body^+(r)\cap F=\emptyset$, $\K(\body^-(r))\cap T=\emptyset$, and $\{ a, \neg b\}\cup  {\OB}_{{\cO}, T}$ is consistent for each $\K b \in F$, there exists an \MKNF rule $r^*\in R$ such that $\body^+(r^*)\cap U\neq\emptyset$.

Note that, since such a set $R$ always exists, so does such an \MKNF rule $r^*$. However, from the proof for (a), there does not exist such an \MKNF rule $r^*$. Thus Case (b) is impossible.

So for each unfounded set $U$ of ${\cK}$ \wrt $(T, F)$, $U\subseteq {\KA}({\cK})\setminus {\atmost}_{{\cK}}(T, F)$.

From (1) and (2), $U_{{\cK}} = {\KA}({\cK}) \setminus \atmost_{{\cK}}(T, F)$.
\end{proof}

\vspace{.1in}
\noindent
{\bf Theorem \ref{them:eq}}\\
Let ${\cK}$ be a normal hybrid \MKNF knowledge base.
{$W_{{\cK}}(\emptyset,  \emptyset)=({\boldP}_\omega,\, {\KA}({\cK})\setminus {\boldN}_\omega)$.}

\begin{proof}
By induction we can prove that $ ({\boldP}_\omega,\, {\KA}({\cK})\setminus {\boldN}_\omega) \sqsubseteq W_{{\cK}}(\emptyset,  \emptyset)$. In the following we show that $W_{{\cK}}(\emptyset,  \emptyset)\sqsubseteq ({\boldP}_\omega,\, {\KA}({\cK})\setminus {\boldN}_\omega)$.

Let $W_{{\cK}}^{(\emptyset, \emptyset)}\!\uparrow^{k}\,= (T_k, F_k)$. 
Clearly, $(T_0, F_0) \sqsubseteq ({\boldP}_\omega,\, {\KA}({\cK})\setminus {\boldN}_\omega)$. Assuming that $(T_i, F_i)\sqsubseteq ({\boldP}_\omega,\, {\KA}({\cK})\setminus {\boldN}_\omega)$, we want to prove that $W_{{\cK}}^{(\emptyset, \emptyset)}(T_i, F_i) \sqsubseteq ({\boldP}_\omega,\, {\KA}({\cK})\setminus {\boldN}_\omega)$.

$T_{{\cK}}^{(\emptyset, \emptyset)}(T_i, F_i) = T^*_{{\cK}, {\KA}({\cK})\setminus F_i}(T_i) \subseteq {\boldP}_\omega$,
{$\atmost_{{\cK}}(T_i, \emptyset) = \Gamma'_{{\cK}}(T_i)$} $ \supseteq \Gamma'_{{\cK}}({\boldP}_\omega)$. By induction, we can assume that $\atmost_{{\cK}}(T_k, F_j) \supseteq \Gamma'_{{\cK}}({\boldP}_\omega)$ for each $0\leq k\leq i$ and $0\leq j< i$. We want to prove that $\atmost_{{\cK}}(T_i, F_i)\supseteq \Gamma'_{{\cK}}({\boldP}_\omega)$.

Let $F_i = {\KA}({\cK})\setminus \atmost_{{\cK}}(T_{i-1}, F_{i-1})$. If $\Gamma'_{{\cK}}({\boldP}_\omega) \not\subseteq \atmost_{{\cK}}(T_i, F_i)$, then there are two possible cases.

Case 1: There exists $r\in {\cP}$ such that $\body^+(r)\subseteq \Gamma'_{{\cK}}({\boldP}_\omega)$ and $\body^+(r) \cap F_i \neq\emptyset$. Then $\Gamma'_{{\cK}}({\boldP}_\omega) \cap ({\KA}({\cK})\setminus \atmost_{{\cK}}(T_{i-1}, F_{i-1})) \neq\emptyset$, thus $\Gamma'_{{\cK}}({\boldP}_\omega)\not\subseteq \atmost_{{\cK}}(T_{i-1}, F_{i-1})$, which conflicts to the assumption for the induction. So this case is impossible.

Case 2: There exists $\K a\in {\KA}({\cK})$ such that $\K a\in \Gamma'_{{\cK}}({\boldP}_\omega)$, $\K a\notin \atmost_{{\cK}}(T_i, F_i)$, $\{a, \neg b\}\cup  {\OB}_{{\cO}, T_i}$ for some $\K b\in F_i$ is inconsistent, and $\{a\}\cup  {\OB}_{{\cO}, T_i}$ is consistent. Then $ {\OB}_{{\cO}, T_i}\models a\imply b$, thus $\K b\in \Gamma'_{{\cK}}({\boldP}_\omega)$. $\K b\in F_i$ implies $\K b\notin \atmost_{{\cK}}(T_{i-1}, F_{i-1})$. Then $\Gamma'_{{\cK}}({\boldP}_\omega)\not\subseteq \atmost_{{\cK}}(T_{i-1}, F_{i-1})$, which conflicts to the assumption for the induction. So this case is also impossible.

Then it is impossible that $\Gamma'_{{\cK}}({\boldP}_\omega) \not\subseteq \atmost_{{\cK}}(T_i, F_i)$. So $\Gamma'_{{\cK}}({\boldP}_\omega) \subseteq \atmost_{{\cK}}(T_i, F_i)$ and $U_{{\cK}}(T_i, F_i) \subseteq {\KA}({\cK})\setminus {\boldN}_\omega$.
So $W_{{\cK}}^{(\emptyset, \emptyset)}(T_i, F_i) \sqsubseteq ({\boldP}_\omega,\, {\KA}({\cK})\setminus {\boldN}_\omega)$ and $W_{{\cK}}(\emptyset,  \emptyset)\sqsubseteq ({\boldP}_\omega,\, {\KA}({\cK})\setminus {\boldN}_\omega)$.
\end{proof}

\vspace{.1in}
\noindent
{\bf Theorem \ref{3.3}}\\
Let ${\cK}$ be a normal hybrid \MKNF knowledge base and $(T, F)$ a partial partition of ${\KA}({\cK})$.
$({\boldP}^{(T, F)}_i,\, {\KA}({\cK})\setminus {\boldN}^{(T, F)}_i) \sqsubseteq W_{{\cK}}(T,  F) \sqsubseteq E_{{\cK}}(T,  F)$, for each $i>0$.

\begin{proof}
 Let $W_{{\cK}}(T,  F) = (T^*, F^*)$. 
We start with 
$({\boldP}^{(T, F)}_0, {\KA}({\cK})\setminus {\boldN}^{(T, F)}_0) = (T, F)$.
It can be verified that $({\boldP}^{(T, F)}_1, {\KA}({\cK})\setminus {\boldN}^{(T, F)}_1) \sqsubseteq (T^*, F^*)$.
Assuming that $({\boldP}^{(T, F)}_i, {\KA}({\cK})\setminus {\boldN}^{(T, F)}_i) \sqsubseteq (T^*, F^*)$ ($i>0$), we want to prove that $(\Gamma_{{\cK}}( {\boldN}^{(T, F)}_i),\, {\KA}({\cK})\setminus\Gamma'_{{\cK}}({\boldP}^{(T, F)}_i)) \sqsubseteq (T^*, F^*)$, which can be similarly proved by the proof for Theorem~\ref{them:eq}.
So $({\boldP}^{(T, F)}_i,\, {\KA}({\cK})\setminus {\boldN}^{(T, F)}_i) \sqsubseteq W_{{\cK}}(T,  F)$, for each $i>0$.
\end{proof}

\vspace{.1in}
\noindent
{\bf Theorem \ref{simplify}}\\
Let ${\cK}$ be a normal hybrid \MKNF knowledge base and $(T, F)$ the well-founded partition of ${\cK}$. An \MKNF interpretation $M$ is an \MKNF model of ${\cK}$ if{f} $M$ is an \MKNF model of ${\cK}^{(T, F)}$.

\begin{proof}
We use ${\cal M}({\cK})$ to be the set of all \MKNF models of ${\cK}$. 
Assuming that for a partial partition $(T', F')$ of ${\KA}({\cK})$, ${\cal M}({\cK}) = {\cal M}({\cK}^{(T', F')})$ and for each $M\in {\cal M}({\cK})$, $M\models_{\MKNF} \bigwedge_{K a\in T'} \K a \land \bigwedge_{K b\in F'} \neg \K b$. We want to prove that ${\cal M}({\cK}) = {\cal M}({\cK}^{W_{\cK}^{(\emptyset, \emptyset)}(T', F')})$ and for each $M\in {\cal M}({\cK})$, $M\models_{\MKNF} \bigwedge_{K a\in T_{\cK}^{(\emptyset, \emptyset)}(T', F')} \K a \land \bigwedge_{\K b\in U_{\cK}^{(\emptyset, \emptyset)}(T', F')} \neg \K b$.

From Proposition~\ref{prop:2}, it is easy to verify that, for each $M\in {\cal M({\cK})}$, $M\models_{\MKNF} \bigwedge_{K a\in T_{\cK}^{(\emptyset, \emptyset)}(T', F')} \K a \land \bigwedge_{\K b\in U_{\cK}^{(\emptyset, \emptyset)}(T', F')} \neg \K b$. Then ${\cal M}({\cK}) = {\cal M}({\cK}^{W_{\cK}^{(\emptyset, \emptyset)}(T', F')})$.

Then the theorem can be proved from the fact that the well-founded partition is equivalent to $W_{\cK}^{(\emptyset, \emptyset)}\!\uparrow^\infty$.
\end{proof}

\end{document}